\pdfoutput=1

\documentclass[11pt]{article}

\usepackage{acl}

\usepackage{times}
\usepackage{latexsym}

\usepackage{xcolor}

\usepackage{amsmath,amsfonts,amsthm} 
\usepackage{wrapfig}
\usepackage{graphicx}
\usepackage{float}
\usepackage{subfig}
\usepackage{cleveref}

\usepackage[T1]{fontenc}

\usepackage[utf8]{inputenc}

\usepackage{microtype}

\title{Blackbird's language matrices (BLMs): a new benchmark to investigate 
disentangled generalisation in neural networks}

 \author{Paola Merlo \and Aixiu An{\small*} \and Maria A. Rodriguez\thanks{*Equal contribution.}\\
         University of Geneva\\
         \texttt{ \{Paola.Merlo, Aixiu.An, Maria.AnduezaRodriguez\}@unige.ch} }

\begin{document}

\maketitle


\begin{abstract}
Current successes  of   machine  learning architectures are based on computationally expensive algorithms and prohibitively large amounts of data.
We need  to develop tasks and data to train networks to reach more complex and more compositional skills.
In this paper, we illustrate Blackbird's language matrices (BLMs), a novel grammatical dataset developed to test a linguistic variant of Raven's progressive matrices, an intelligence test usually based on visual stimuli.
The dataset consists of  44800 sentences, generatively constructed to support investigations of current models' linguistic mastery of grammatical agreement rules and their ability to generalise them.
We present the logic of the dataset, the method to automatically construct data on a large scale and the architecture to learn them. Through  error analysis and several experiments on variations of the dataset, we demonstrate that this language task and the data that instantiate it provide a new challenging testbed to understand generalisation and abstraction.
  
\end{abstract}

\section{Introduction}

The current reported success of machine learning architectures is based on computationally expensive algorithms and prohibitively large amounts of data that are available for only a few, non-representative languages. Moreover, despite these large training data requirements, current architectures still are not able to understand new sentences or derive the meaning of a word never seen before.
To reach better, possibly human-like, abilities in neural networks' abstraction and generalisation, we need  to develop tasks and data that help us understand their current generalisation abilities and help us train them to more complex and compositional skills.

 Generalisation in NLP has been defined in a very narrow way, as extension from a set of data points to new data points of exactly the same nature (i.i.d. assumption). Not much effort has gone in trying to generalise to new problems or out of distribution \citep{scholkopf2019}.
Even under this very narrow definition, recent studies show that current algorithms 
do not generalise well \citep{belinkov-bisk2018,belinkov-glass2019tacl}.

 One likely reason why people generalise better is that they have a strong prior bias, grounded in the actual structure of the problem.
  A large body of literature of experimental work has demonstrated that the human mind is predisposed to extract regularities and generate rules from data, in a way that is distinct from the patterns of activation of neural networks \citep{sable-meyer-ea2021}.  
One possible approach to develop more robust methods, then, is to pay more attention to the decomposition of complex observations, discovering the factors in the generative process that gives rise to the data \cite{scholkopf-etal2012}. 
 To study how to discover the underlying problem structure, machine learning research in vision has developed  the notion of disentanglement.
A disentangled representation can be defined as one where single latent units are sensitive to changes in single generative factors, while being relatively invariant to changes in other factors \cite{bengio-ea2013}. 

To  learn more disentangled linguistic representations, that reflect the underlying linguistic rules of grammar, we develop a new linguistic task.
We will use a new set of progressive matrices tasks developed specifically for our goals and demonstrate their usefulness.
While the use of automatically generated data in NLP is not new, this kind of progressive matrix generation for higher level linguistic reasoning has never been tried before for NLP,  as far as we are aware.\footnote{The code and the data described in this paper will be released in full on publication. Currently, more examples of the data and more details on models specifications are to be found in the supplementary materials.}

\section{\hspace{-.2cm}Blackbird's Language Matrices (BLMs)}

Inspired by computational methods on vision,  we develop a new linguistic task, to  learn more disentangled linguistic representations that reflect the underlying linguistic rules of grammar.

The solution of the tasks requires identifying the underlying rules that generate compositional datasets  (like Raven's progressive matrices), but for language. We call them Blackbird's Language Matrices (BLMs).

\subsection{Progressive matrices for visual stimuli}

\begin{figure}
  \center  \includegraphics[width=0.8\linewidth]{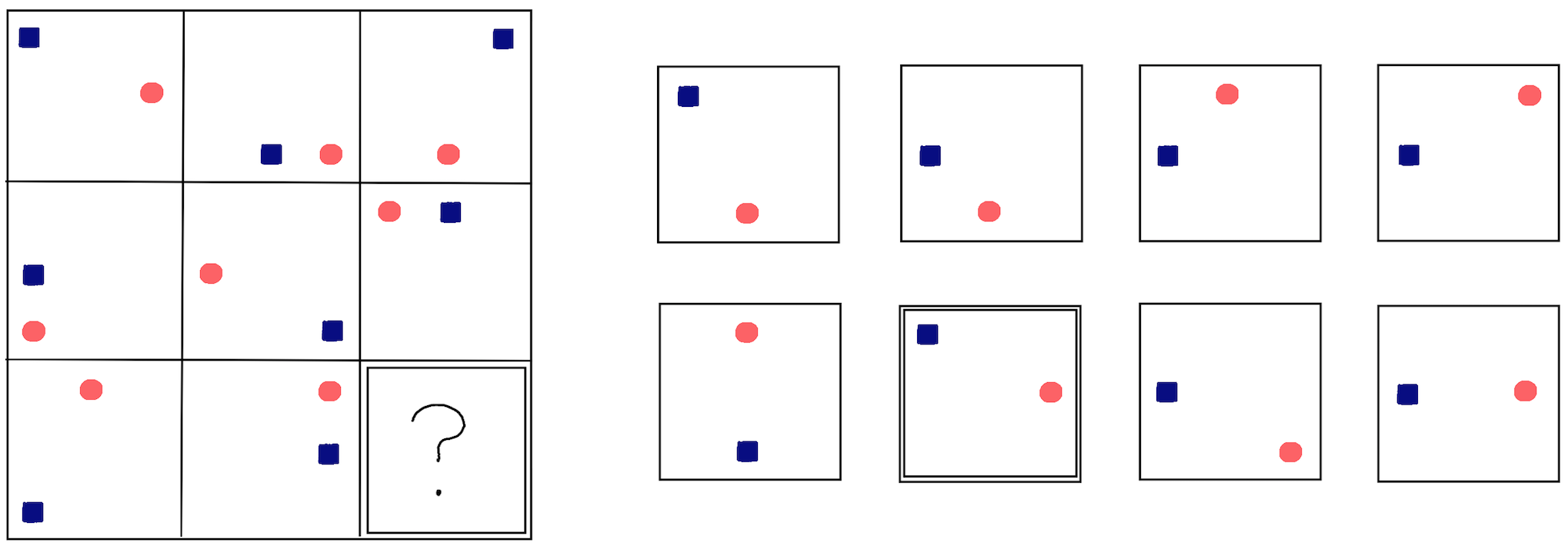}
\caption{Examples of progressive matrices in the visual world. The task is to determine the missing element in a visual pattern:  Given the matrix on the left, choose the last element of the matrix from the choice of elements on the right.
The matrix is constructed according to two rules:
Rule 1: from left to right, the red dot shape moves one place clockwise each time. This pattern continues onto the next row; Rule 2: from top to bottom, the blue square moves one place anticlockwise each time. This pattern continues onto the next column. Identifying these rules leads to the correct answer: the correct answer is marked by double edges; it is the only cell that continues the generative rules correctly.}
\label{RPMtest2-3}
\end{figure}

Raven's progressive matrices (progressive because tasks get harder) are IQ tests consisting of a sequence of images (usually eight) connected in a logical sequence by underlying generative rules.  
The task is to determine the missing element (usually the last) in this visual sequence. An example and explanation of this task is given in Figure \ref{RPMtest2-3}. The matrices are built according to  generative rules that span the whole sequence of stimuli and the answers are constructed to be similar enough that the solution can be found only if the rules are identified correctly.

Traditionally, progressive matrices  as intelligent tests are designed by hand, but recent research in vision that has used this task to train neural networks has typically employed some structured generative model to create larger numbers of questions \citep{wang-su2015}.
In this way, a correct answer  is consistent with the underlying generative model, so the learning process basically discovers how to induce the model. In this way, for example, it is possible to identify clear dimensions of successful and unsuccessful generalisation. For example, matrices for vision have shown that the best models can  apply known abstract relationships in novel combinations, but fail in applying known abstract relationships to unfamiliar entities  \citep{barrett-ea2018}.

\subsection{Progressive matrices for language}

Reaching  the right answer in an RPM requires solving several subproblems (Carpenter et al 1990):
one has to  identify the abstract structure of the matrix, identify the elements  manipulated by the rules in the matrix, the relevant attributes of these elements and the rules of change of these attributes.

To instantiate these subproblems in the automatic generation of language matrices, first, we define the linguistic phenomenon that needs to be learnt.
Second, we define the rules governing the abstract automatic generation process that embody the properties of the linguistic phenomenon and  we compose rules into grammatical language templates.
Finally, we automatically create large samples of data with great  variety so that the underlying structure of the problems needs to be discovered.
We describe these steps below. 

In describing the process to build BLMs, we will talk of \textit{contexts}, the sequence of sentences whose last element needs to be identified, and \textit{answer set}, the set of answers that instantiates the multiple choice task that needs to be solved by the BLM test.
An example of the resulting  context and answer set is shown in Figure~\ref{BLM-agreement}.

\begin{figure*}
\centering
\begin{tabular}{llll} 	\hline
\multicolumn{4}{c}{Template for contexts} \\	\hline
1 Subj-sing & N1-sing & & Verb-sing\\
2 Subj-plur & N1-sing & & Verb-plur\\
3 Subj-sing & N1-plur & & Verb-sing\\
4 Subj-plur & N1-plur & & Verb-plur\\
5 Subj-sing & N1-sing & N2-sing & Verb-sing\\
6 Subj-plur & N1-sing & N2-sing & Verb-plur\\
7 Subj-sing & N1-plur & N2-sing & Verb-sing\\
8 Subj-plur & N1-plur & N2-sing & Verb-plur\\\hline
\end{tabular}
\vfill
\begin{tabular}{llll} 
\hline
\multicolumn{4}{c}{Example of  contexts}\\
	\hline
	1  The computer   & with the program  &  & is broken. \\
	2 The computers   & with the program  &  & are broken.\\
	3 The computer    & with the programs &  & is broken. \\
	4 The computers   & with the programs &  & are broken.\\
	5 The computer    & with the program  & of the experiment & is broken. \\
	6 The computers   & with the program  & of the experiment & are broken.\\
	7 The computer    & with the programs & of the experiment & is broken. \\
	8  ???\\
\end{tabular}
\vfill
\begin{tabular}{ll} 
\hline
\multicolumn{2}{c}{Example of  answers}\\
	\hline
 1 The computer with the program  and the experiment  is broken.  & Coord\\
 2 \textbf{The computers with the programs  of the experiment  are broken.} & Correct\\
 3 The computer with the program    is broken. & WNA\\
 4 The computer with the program  of the experiment  are broken. & AE\\
 5 The computers with the programs  of the experiment  are broken. &   Alter N1\\
 6 The computer with the programs  of the experiments  are broken. & Alter N2\\
\hline 
\end{tabular}
\caption{Examples of template and progressive matrices for agreement. The examples here are shown in English for ease of exposition. The original data was developed for French. The sequence is generated by a rule of progression of number of attractors (one and two), a rule of subject-verb agreement that alternates every sentence between singular and plural of the head noun and a rule of number of the first attractor that alternates between singular and plural every two sentences, and a second attractor whose number does nor vary (it is always singular). Thus, the correct answer for this example is a sentence that has three noun phrases and a plural subject and plural first attractor and singular second attractor. There are three types of context: main clause, completive clause, and relative clause. We present here an example main clause. 
The answer set is a multiple choice of the correct answer among other incorrect alternatives, illustrated here: Coord: the last noun appears in a coordination phrase; WNA: Wrong number of attractors; AE: Agreement error;  Alter N1: wrong number of the first NP attractor; Alter N2: wrong number of the second NP attractor.}
    \label{BLM-agreement}
\end{figure*}

\subsubsection{Choosing the body of grammatical rules}

We choose to construct data to determine if the rules of subject-verb number agreement can be learned.
As a reminder, the main rule of subject-verb agreement in French, and English, states that subject and verbs agree in their number, and they do so independently of how many noun phrases intervene between the subject and the verb, as shown in the context examples of Figure \ref{BLM-agreement}.\footnote{In practice, the intervening noun phrases can act as agreement attractors and trigger agreement mistakes, if they are close to the verb, like for example the fourth sentence in the answers in Figure \ref{BLM-agreement}.}

Subject-verb agreement is a morphological phenomenon of appropriate complexity to start our investigations with BLMs. Subject-verb agreement is clearly limited to some specific words in the sentence, so that the elements and the attributes manipulated by the underlying rules can be clearly identified. It is marked explicitly in the forms of words (for example by an $-s$ ending) and it does not depend on the words' meaning.  Moreover,  agreement rules show structural properties, so that sequences of increasing complexity of application of the rule can be defined \cite{linzen-ea16,linzen-leonard18}.
We choose to work specifically on French because its agreement system, its verb conjugations and its noun phrase structure lends itself well to our investigation.

\subsection{Creating the BLM templates}

At an abstract level, the structure of each matrix is defined as a triple (or a set of triples) of relation type R
(e.g.progression or alternation, among others), object types O, such as the nonterminals of a grammar, 
and attributes of the objects, A. 

\paragraph{\sc Creating the  templates contexts }
For the (semi)-automatic generation of the contexts, first, we define the abstract structure of the matrix as a progression in the number of attractors, and alternations of the  number of the noun phrases, as shown in the templates of Figure \ref{BLM-agreement}. The `objects' that we manipulate are noun phrases and we manipulate their agreement attributes: the number of the head noun (which needs to match the number of the verb), the number of the closest noun and the  number of the second noun (both can vary freely), as in Figure \ref{BLM-agreement}.

So, for example, the sequence in Figure \ref{BLM-agreement} is generated by a rule of progression of number of attractors (one and two), a rule of subject-verb agreement that alternates for every sentence between singular and plural of the head noun, a rule of number of the first attractor that alternates between singular and plural every two sentences, and a second attractor whose number does not vary (it is always singular). Thus, the correct answer for this example is the sentence indicated in bold in the answer set.

 The data generation process can be formalised by an attribute-value grammar that defines the structure of the matrix, as shown in section \ref{fig:template-grammar}, in the appendix of supplementary materials \ref{sec:appendix}.
This grammar defines the template sentence structures in which the  manipulated elements are embedded  (main clause, completive, relative), the internal structure of the sentences and the attributes of the nonterminal objects.
The constraints applied to the grammar (indicated as the rules) create the curated structure of the matrix.

\paragraph{\sc Creating the template answers}

The generation of possible answer sets is also a complex issue. Alternative (and incorrect) answers need to be sufficiently distinguishable. The correct answer cannot be found by just learning some of the generative rules, instead of all of them, or by some simple heuristic. 
We define a fixed set of alternatives, which belong to predefined constructions belonging to very precise minimal pairs compared to the correct answer, that support fine-grained error analysis. We opt for a choice among grammatical and ungrammatical alternatives comprising the  choices exemplified in Figure \ref{BLM-agreement}, illustrated for the main clause structural context. Across matrices, the order of presentation of the different types of answers is rotated for all possible positions of all the possible answers, to avoid a positional identification of the correct answer. See also the original French alternatives in Figure \ref{answers} in the appendix.

\subsubsection{Creating the data sample and lexically-varied variants}

\begin{figure*}
\footnotesize
\begin{tabular}{p{0.48\linewidth}p{0.48\linewidth}}
\hline
\hline Example & Translation\\\hline
1 L'expérience   avec la peinture   a rencontré un grand succès. & \textit{The experience with the painting has met with great success.}\\
2 Les travaux   avec la peinture   ont rencontré un grand succès. & \textit{Works with the painting have met with great success.}\\
3 L'association   avec les peintures   a rencontré un grand succès. &\textit{ The association with the paintings has met with great success.}\\
4 Les séances   avec les peintures   ont rencontré un grand succès. & \textit{The sessions with the paintings have met with great success.}\\
5 L'activité   avec la peinture   de l'enfant a rencontré un grand succès. & \textit{The activity with the painting of the child has met with great success.}\\
6 Les créations   avec la peinture   de l'enfant ont rencontré un grand succès. & \textit{The creations with the painting of the child have met with great success.}\\
7 L'activité   avec les peintures   de l'enfant a rencontré un grand succès. & \textit{The activity with the paintings of the child has met with great success.}\\
\end{tabular}
\caption{Matrix context of type II for the main clause context. Translations from French are as literal as possible for expository purposes.}
\label{BLM-typeII}
\end{figure*}

Once the BLM templates are defined, we create a large sample of natural, grammatical sentences, to train the networks and to generate the answer sets.

\paragraph{Creating natural contexts and answers}
We use a semi-automatic method, based on context-aware word embeddings, specifically we use Camembert \citep{martin-etal-2020-camembert} to generate more sentences. 
The general process consists in masking some of the nouns to generate other  most probable nouns and use them to construct new sentences.
More specifically, we mask different nouns in the three kinds of constructions. 
Section \ref{masking-examples} in the appendix shows how the noun phrases were masked.

\paragraph{Creating three types of natural contexts and answers}

The semi-automatic augmentation step described above can be used to create increasing levels of lexical variation inside the same matrix, intuitively creating progressively harder and harder problems. 

Type I matrices have the same lexical items in each matrix, as shown in the example in Figure 2. Type II matrices introduce partial lexical variation, with variation for one noun phrase at a time. They are illustrated in Figure \ref{BLM-typeII}. In Type III, sentences that instantiate the same values in the template are shuffled and matrices with full lexical variation are created. Figure \ref{lexically-varied-contexts} shows an example of a  matrix with lexically varied sentences.  
We generate lexical variants of answers by the same procedure used in the contexts.

\begin{figure*}
\footnotesize
\begin{tabular}{p{0.48\linewidth}p{0.48\linewidth}}
\hline
\hline\textbf{Contexts} \\\hline Example & Translation\\\hline
1  La conférence  sur l’histoire  a commencé plus tard que prévu. & \textit{The talk on history has started later than expected.}\\
2 Les responsables   du droit vont démissionner. & \textit{Those responsible for the right will resign.}\\
3    L’ exposition  avec les peintures  a rencontré un grand succès. & \textit{The show with the paintings has met with great success.}\\
4    Les menaces  de  les réformes  inquiètent les médecins. & \textit{The threats of reforms worry the doctors.}\\
5   Le trousseau  avec la clé de la cellule repose sur l’étagère.& \textit{The bunch of keys of the cell sits on the shelf.}\\
6   Les études  sur l’effet de la drogue apparaîtront bientôt. & \textit{The studies on the effect of the drug will appear soon.}\\
7    La menace  des réformes  dans l’ école inquiète les médecins. & \textit{The threat of reforms in the school worries the doctors.} \\\hline
\textbf{Answers}\\\hline
Example & Translation \\\hline
1  Les nappes  sur les tables et le banquet brillent au soleil. & \textit{The tablecloths on the table and the console shine in the sun.}\\
2 \textbf{Les copines   des propriétaires de la villa dormaient sur la plage.} & \textit{The friends of the owners of the villa were sleeping on the beach.}\\
3  Les avocats   des assassins  vont revenir. &\textit{The laywers of the murderers will come back.}\\
4  Les avocats   des assassins  du village va revenir. &\textit{The lawyers of the murderers of the village will come back.}\\
5  La visite   aux palais  de l’ artisanat approchent. & \textit{The visit of the palace of the crafts is approaching.}\\
6  Les ordinateurs avec le programme des expériences sont en panne. & \textit{The computers with the program of the experiments are broken.}
\end{tabular}
\caption{Example of lexically varied contexts for the main clause contexts.  Correct answer in bold. 
}
\label{lexically-varied-contexts}
\end{figure*}

With these novel datasets, we train learners and study their ability to learn the underlying rules giving rise to subject-verb agreement. As illustrated by Figure \ref{lexically-varied-contexts}, this task can get quite difficult. We explore whether current models can learn to perform the task at all, and what factors of variation help the learning of the underlying rules.

\section{Learning the matrices}

We train several models to investigate whether we can learn to solve this task at all and, more specifically, how well we can learn the task through  a latent disentangled representation of the matrix.
The computational choices of the problem concern the representation of the data set, the representation of the actual sentences and sequence of sentences and, finally, the computational architecture. We describe these methodological components below.

\paragraph{Data and embeddings}

The training data  consists of  44800 BLMs (sequences of 7 context sentences and the corresponding correct continuation), split into 80\% for training (35480) and 10\% for validation (4480). For testing, we have  4480 sequences of 7 sentences as BLMs and  6 possible answers for each sequence. 
To obtain representations of our data, we use {\sc FlauBERT}, a transformer model for French \citep{le-etal-2020-flaubert-unsupervised}, pretrained using a masked language modeling objective similar to BERT \citep{Devlin:2019gm}.
As this model gives us representations in context for each word in a sentence, we create an average representation for our sentences, so that we have a single vector for the entire sentence.\footnote{In these averaged representations, we omit the special tokens (e.g. $\left[ CLS \right ]$, $\left[ SEP \right ]$) added by the transformer at the beginning and end of a sentence. In addition, we pad each sentence to have the same length as the longest sentence in the data. }

\subsection{Our model}

\begin{figure*}
\includegraphics[width=\linewidth]{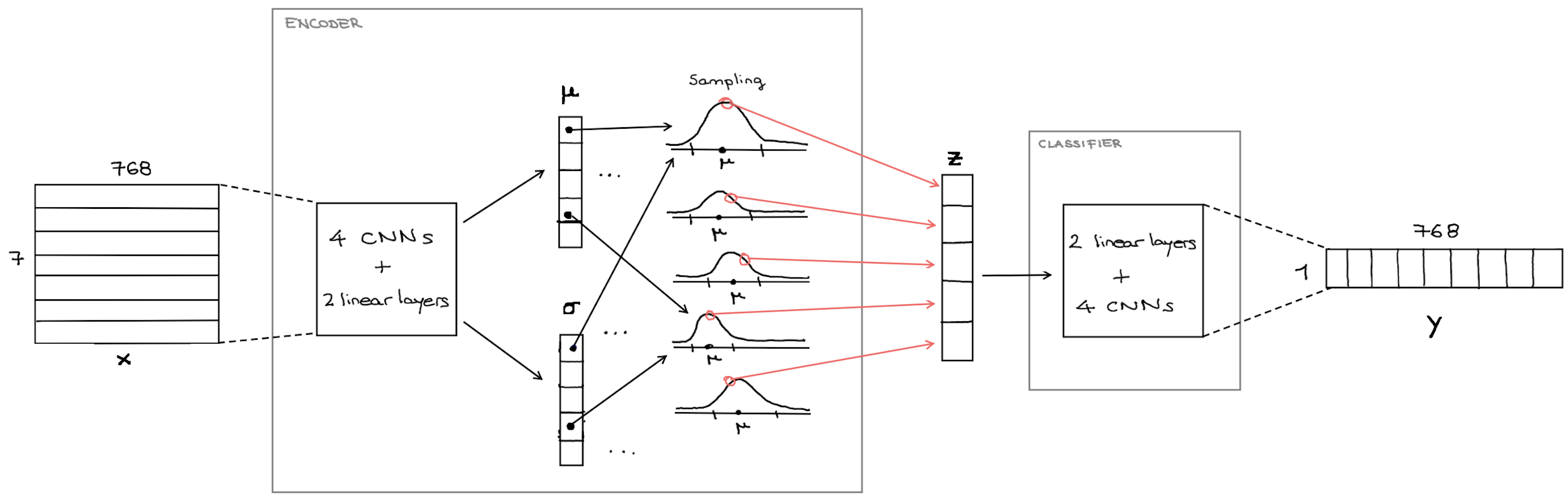} 
\caption{Model architecture} 
\label{fig:model-architecture}
\end{figure*}

We develop a variational information bottleneck (VIB) approach \citep{alemi-ea2017}, to model both the fact that the generative process that gives rise to the data is combinatorial and structured, and also that the context sentences need to be learnt as a logical sequence. A picture of the architecture is shown in Figure \ref{fig:model-architecture}. 

In our encoder-decoder approach, instead of reconstructing the original input (the sequence of sentences) like in a VAE, the decoder produces a new and compressed representation of the input. The intuition is that the latent layer, having learned the important factors of the input, can be used as is to produce a new single vector which represents the sequence and is close to the answer. The new vector will then be compared, at training time, with the correct answer. The model is updated after an entire sequence of seven sentences and the loss function is computed between the produced vector and the true answer vector.

\paragraph{Encoder and decoding classifier }

The encoder is composed of four one-dimensional CNNs, three fully connected layers and ReLU activation.%
\footnote{Structure and dimensionality: Input: 1x768x7 ; 
Encoder: four CNNs 1x100x7 (stride 1), three fully-connected layers 1x300; Output: 1x10 (dimensionality of the latent vector).}
The latent vector $z$ is of size 5. 
The structure of the decoding classifier is the mirror image of  the encoder, but with a different output dimensionality, as shown in Figure \ref{fig:model-architecture}. It is composed of three fully connected layers, four one-dimensional  transposed CNNs and ReLU activation.%
\footnote{Structure and dimensionality:
Input: 1x10 (latent vector); 
Encoder: three Fully connected layers of dim 1x300, four Transposed-1d-CNN of dim 1x100x1 (stride 1);  Output: 1x768x1 (new representation of the input).}

\paragraph{Loss function}

Because our goal is to learn the generative factors of the data in a disentangled representation, the  $\beta$-VAE objective function is used. The $\beta$-VAE loss is composed of the $\beta$ coefficient and two different terms: the reconstruction loss and the Kullback-Leibler divergence \cite{higgins2016early}. 

The reconstruction loss is the binary cross-entropy between the output of the model and the true answer. 
The $D_{KL}$ regulariser constraint is added to the loss function to force closeness to a prior and embodies pressure for redundancy reduction and latent factors independence.  Because the prior is a unit Gaussian, this factor can be increased in importance by the use of a simple coefficient, to learn statistically independent factors, that is, a disentangled representation. 
\citet{Higgins2017betaVAELB} propose the addition of a single hyperparameter $\beta$ to the original framework to limit the capacity of the latent variable and control the rate of learning of independent factors.  A coefficient of $\beta = 1$ corresponds to the original loss function, while  $\beta > 1$  pushes to learning more disentangled latent representations. 
%

%

Equation \ref{objFunctBVAE} 
shows our objective function, where $x$ is our input, $z$ is the latent variable, $\beta$ is our disentangled hyperparameter and $D_{KL}$, the Kullback-Leibler divergence, is the regulariser that forces the solution to be close to the prior, and $p(z)$ is a Gaussian prior, $q_\phi$ refers to the encoder and $p_\theta$ refers to the decoder, and  $L_{BCE}$ is binary cross-entropy loss.
Notice that, in our model, $x$ and $y$ are not the same, as $y$ is not the reconstruction of $x$. Instead, the output $y$ is a vector of real numbers. 
In binary cross entropy ---the loss function we use--- these numbers are interpreted as a distribution over $\{0,1\}$.

\begin{equation} \label{objFunctBVAE}
\nonumber
\begin{split} 
& (1) \hspace{0.5cm}\mathcal{L}(\theta, \phi; x, z, \beta) = \\ 
&\mathbb{E}_{q_\phi(z|x)} [ L_{BCE}(f_{\theta}(z), y) ] - \beta D_{KL}(q_\phi (z|x) \big\|p(z))
\end{split}
\end{equation}

\section{Experimental results}

We perform several experiments to investigate how the properties of the dataset support the learning of subject-verb agreement and what latent representations are developed.%
\footnote{
Beside the reported experiments, we also investigate if the sentence type (main, completive or relative clause) has an effect on the results and errors. In terms of correct answers, main clauses are actually the hardest ones to learn. For all the three datasets, we observe roughly the same distributions in terms of correct answers and types of errors. In the reported analysis of results and errors,  then we no longer distinguish by sentence type.
}

\paragraph{Experimental results by data type and $\beta$ value}

\begin{figure*}
\scriptsize
\centering
\includegraphics[width=\linewidth]{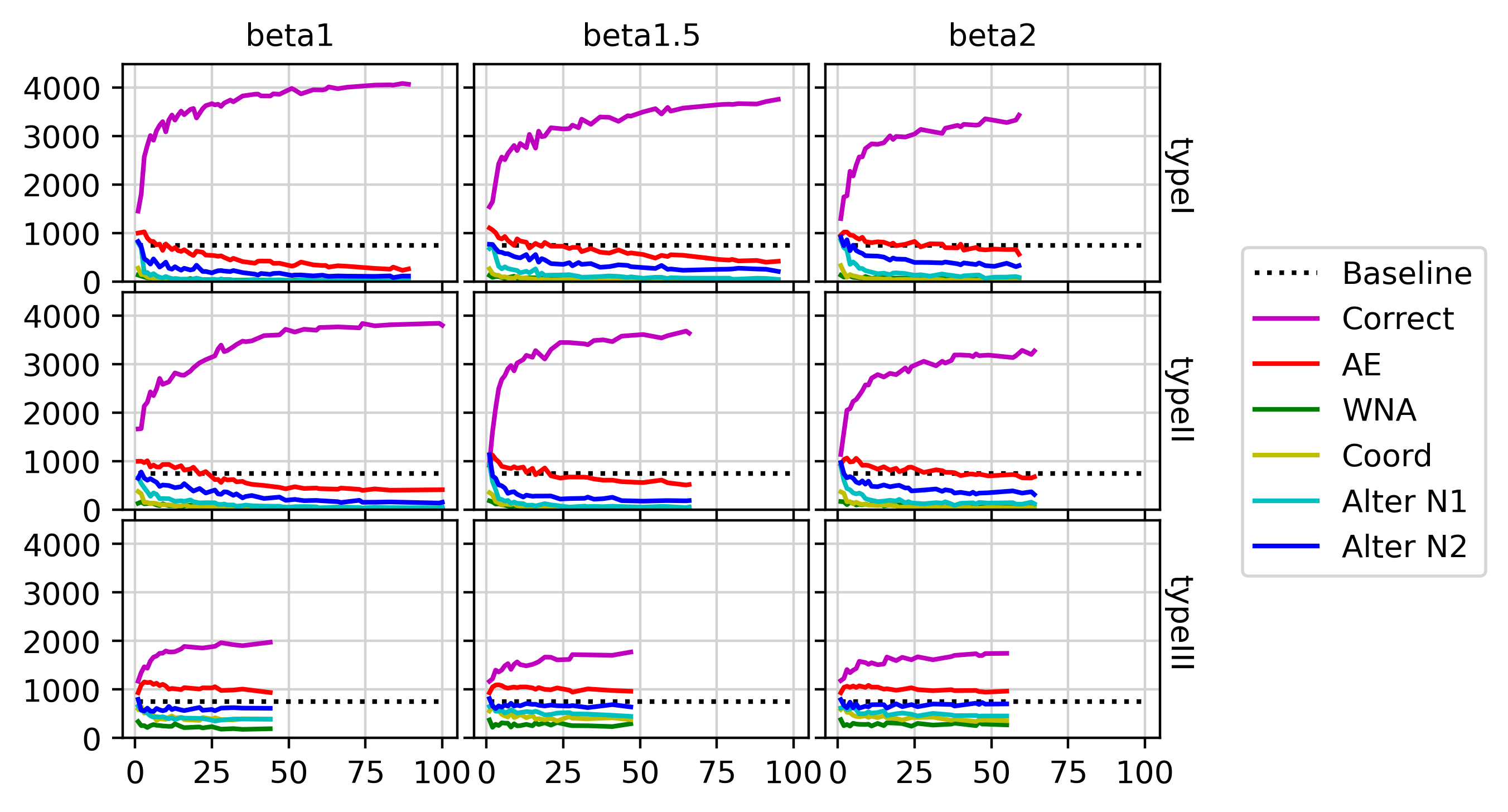} 

\caption{Results on test set for the  models.}   
\label{fig:results}
\end{figure*}

Figure \ref{fig:results} presents results for the three data sets: the basic data set (type I), the partially lexically-varied data set (type II) and the fully lexically-varied data set (type III). 

For the more successful configurations,  we can observe, first of all, that the learning curve is steep, especially for type I, showing already a good rate of learning after a few epochs, reaching a good 84.8\% accuracy in the best, and easiest, case. $\beta = 1$ yields the best results ($\beta = 1$ is the value of a normal VIB that does not force disentanglement). This result then confirms what already found in the vision literature that entangled representations lead to better accuracy \citep{Higgins2017betaVAELB}.

If we class the data to train the models along a gradient of  lexical variation, from  
least in type I to most in type III, we can see that task accuracy is proportional to the amount of lexical variation, the more the harder.
This indicates that  lexical variation does not help identifying the underlying formal invariants of the examples. This latter result is analogous to findings in vision that the best models can  apply known abstract relationships in novel combinations, but fail in applying known abstract relationships to unfamiliar entities \citep{barrett-ea2018}.

\paragraph{Error analysis}
Figure \ref{fig:results} also shows interesting patterns of errors, which are consistent across  the  three types of datasets. The different errors indicate ability of the models to learn different types of information: subject-verb agreement requires long-distance, structural information; errors on N1 and N2 tell us whether the model exhibits recency effects, thereby showing, like humans, that both structural and linear considerations come into play in learning agreement; choosing the wrong number of attractors is a very salient form of structural deviance from the correct answer and coordination is a more subtle  one. 

For all three types, agreement errors are always the most frequent, followed by N1 and N2 alternatives, while coordination and number of attractors mistakes occur much less frequently, suggesting the models do learn the difference in construction and the rule of attractor sequence. This result matches our intuitions that these are also the two most saliently different cases from the right answer, because they differ in structure.

\paragraph{Analysis of disentanglement by ablation of latent factors}

Recall that disentangled representations are those where the  single latent units are sensitive to changes in single generative factors. It can be  determined whether the latent representations are disentangled by ablation studies that modify portions of the latent vectors. These should give rise to predictable effects, specifically predictable increase of error types based on different generative factors. BLMs contexts are built by generative rules and the  answers set was chosen  according to specific patterns.
So, assuming we have a latent space that encodes the underlying generative rules, we can make the predictions shown in Table \ref{tab:traversal-prediction}.

\begin{table}
    \footnotesize
    \begin{tabular}{p{0.59\linewidth}|p{0.31\linewidth}} \hline
     Rule    &  Expected error\\\hline
    \textbf{(R1) Subj-verb agreement:} The subject and the main verb agree. They alternate between singular and plural at each sentence.    & Increase in agreement errors (AE).\\
   \textbf{(R2)  Number of attractors:} The number of attractors grows with increments of one. & Increase in coordination and number of attractors errors (Coord and WNA).\\
    \textbf{(R3)  N1 grammatical number:} The number of the first attractor alternates between singular and plural in increments of two. & Wrong number of the first attractor NP (alter N1). \\
     \textbf{(R4) N2 grammatical number:} The number of the second attractor never changes. & Wrong number of the second attractor NP (alter N2).\\\hline
    \end{tabular}
    \caption{Predictions of expected errors for masking each latent factors.}
    \label{tab:traversal-prediction}
\end{table}

\begin{figure*}
\centering
\includegraphics[width=\linewidth]{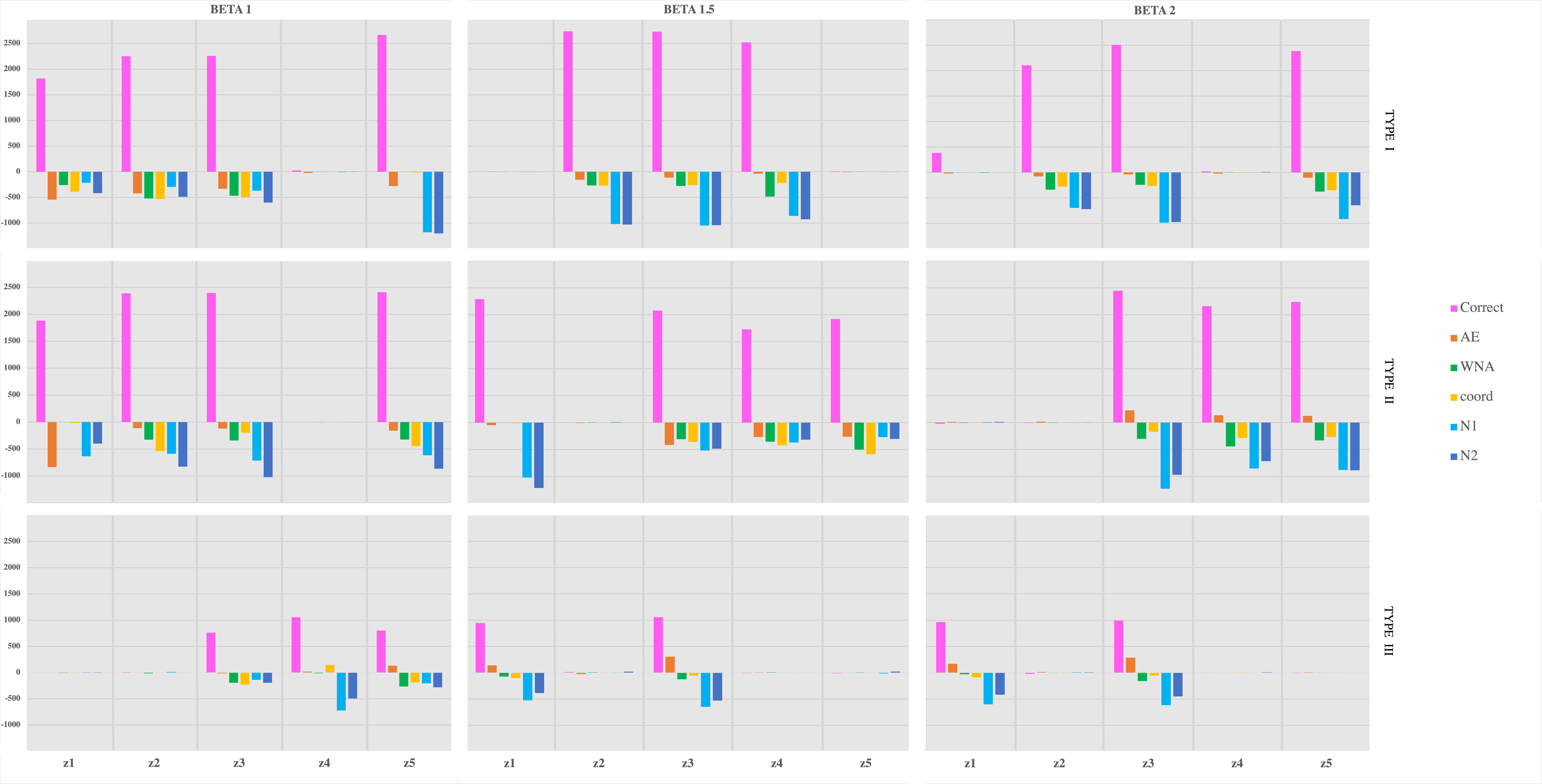} 
\caption{Error analysis by ablation of latent components. The y-axis shows the difference in errors between the non-ablated and the ablated model for each ablated component of the latent vector $z$, when $z=5$.}
\label{fig:latent-ablations}
\end{figure*}

The process to test these expectations in error patterns consists in learning the latent vector in the usual way, then masking one of the latent vector components (putting its value to 0) and running the classifier with this partially-masked vector. Analysis of errors on the test set by ablated latent component are shown in Figure \ref{fig:latent-ablations}, where each latent factor is ablated, for the three different types of data (shown vertically) and different values of $\beta$, horizontally.

We can observe that all ablated models do overall worse than the non-ablated one, and also that, as $\beta$ increases, more factors are not useful. For the errors, no clear pattern emerges. If we look at what kind of error is majoritarily degraded, we can say that agreement errors are markedly degraded if $z[1]$ is masked in $\beta = 1$, and 
$z[5]$ appears to be associated to the rule on number of attractors (R2). For $\beta = 1.5$, $z[1]$ does not encode structural factors and shows an increase of the factors encoding attractors. In general though, there is no clear trend of disentangled factors.

\subsection{Creating data variants: Unordered matrices}

\begin{figure*}[ht] 

\centering
\includegraphics[width=0.7\linewidth]{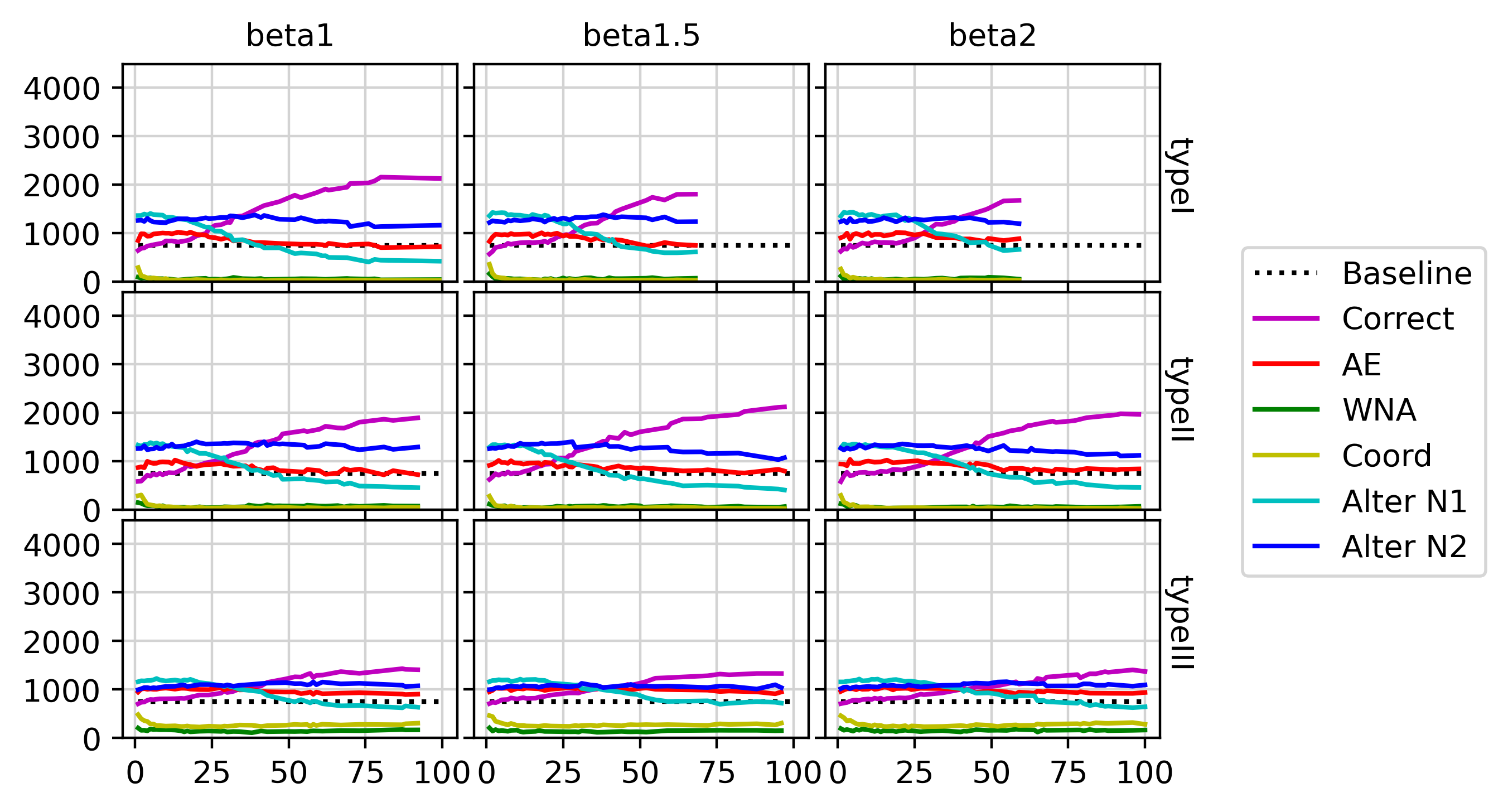}
\caption{Results on test set for the  models trained on unordered shuffled matrices for the three types of datasets. Dotted horizontal random baseline.}   
\label{fig:results-shuf}
\end{figure*}

Natural language tasks and problems often are not limited to a single sentence but span over several sentences, for example, in textual entailment, machine translation, reference resolution, dialogue.  It is therefore important to build test sets that stress language learning abilities and from which we can extrapolate realistic conclusions about real learning patterns.
An interesting and defining property of the data we use is that they constitute logically  connected sequences of sentences, and, as such, the ordered nature of the data is expected to be important in solving the task. 
To test whether the model really exploits the order of the data, we compare the results presented above to results obtained with unordered matrices. We shuffle each matrix, to keep the templates constant, but make the order random. Results are shown in Figure \ref{fig:results-shuf}. They show that the ordered presentation is important in learning the right answer, as the accuracy drops by half. Moreover, quantitatively, while errors in the ordered matrices are never higher than the random baseline, some types of errors do increase above the baseline in unordered matrices. Qualitatively, the results also show that without matrix order information the pattern of mistakes changes: the mistakes that are interpreted as the model giving too much weight to linear recency (N1 and especially N2) become dominant.

\paragraph{Discussion}

The experiments show that BLMs define a hard, but learnable task. Their structured construction lends itself to controlled experiments, and the task is challenging enough that the models make informative patterns of mistakes. Our models can learn BLMs,  show that lexical uniformity affects the accuracy of learning, and that the best results are still at least partially entangled. 
Finding models which learn disentangled representations of this task is a challenging open problem, which can help our understanding of different deep learning architectures.

\section{Related work }

The current paper  does not have any direct comparison, as, to our knowledge, this is the first proposal of a dataset for language using BLMs. 
But it is inspired by and situated among work on disentanglement and generalisation for vision, where RPMs datasets have been used, and it contributes to the investigation  on learning of agreement by neural networks.

\paragraph{Disentanglement datasets for vision and language}

\citet{vansteenkiste2020disentangled} develop a dataset for vision to learn  tasks similar to Raven's Progressive Matrices, and evaluate the usefulness of the representations learned for abstract reasoning tasks.  They observe that disentangled representations enable quicker learning using fewer samples.
RPMs and their language equivalent have not been used before for language, as far as we know, but one other dataset exists to learn disentanglement for language, dSentences \citep{mcharrak2018}.\footnote{This dataset is similar to the dSprites dataset for vision (https://github.com/deepmind/dsprites-dataset))}  Like our  dataset, it is  large in terms of size, but, unlike our dataset, the examples are unrealistically simple. The factorial combinations of very simple sentences with a few morphological marking does not constitute a sufficiently realistic challenge from the linguistic point of view. Moreover,  natural language tasks and problems often span over several sentences, as discussed above, so the ordered sequence that characterises our BLMs is a crucial difference.

\paragraph{Related work on disentanglement for vision and language}

In the literature on disentanglement for vision, 
\citet{higgins2016early} and related work  propose an approach to variational autoencoders  based on  redundancy reductions, and pressure to learn statistically independent factors. The  disentangled representations enable zero-shot learning and emergence of visual concepts \cite{higgins2018scan,burgess2018understanding}. Following work demonstrates that inductive biases are necessary to learn, but can be successfully encoded  in potentially imprecise and incomplete labels \cite{locatello2020disentangling}. 
\citet{mercatali2021disentangling} extend VAEs  to learn discrete representations appropriate for language. The proposed model outperforms continuous and discrete baselines on several qualitative and quantitative disentanglement benchmarks and extrinsic evaluations. 
\citet{zheng-lapata-2022-disentangled} propose a disentangled extension to sequence-to-sequence models which encourage disentanglement by adaptively re-encoding the source input.

\paragraph{Related work on learning agreement}

Previous work on agreement  has tested  recursive neural  network (RNN)
language  models and  found that  RNNs  can learn  to predict  English
subject-verb agreement, if provided with explicit supervision \cite{linzen-ea16}. Follow-up work has shown that  RNNs  are better  at
long-distance agreement  if they  can use  large vocabularies  to form
rich  lexical  representations  to  learn  structural  patterns \citet{bernardy-lappin17}.  
\citet{gulordava-ea18} extends previous work  to  four  languages  of different  linguistic
properties (Italian, English, Hebrew, Russian) and shows the models  make accurate
predictions and compare well with  humans, thereby suggesting that the
networks learn deeper grammatical competence.
 Recent work  by
\citet{lakretz2021mechanisms} studies RNNs in more  detail, looking
at  single   neurons,  and   finds  that  individual   neurons  encode
linguistically meaningful  features very saliently and  propagate  subject-verb
number agreement information over time.

\section{Conclusions}

In this paper, we have introduced  Blackbird's language matrices (BLMs), a novel linguistic dataset, generatively constructed to support investigations in representation learning of grammatical rules. Through  error analysis and several experiments on variations of the dataset, we demonstrate that this language task and the data that instantiate it provide a new testbed to understand generalisation and abstraction.
The larger contribution of the paper lies in the definition of a new challenging task, and the development  of a general procedure to develop many other such datasets, on different linguistic problems. But it also lies in tackling a mixture of language tasks and reasoning to take us closer to investigations of human linguistic intelligence.

\section*{Acknowledgments}
 We gratefully acknowledge the partial support of the NCCR Evolving Language, Swiss NSF Agreement 51NF40\_180888.
\bibliographystyle{acl_natbib}
\bibliography{BLM22}

\newpage
\appendix
\onecolumn
\section{Supplementary Materials}
\label{sec:appendix}

\subsection{Grammar to create matrices automatically}
\label{fig:template-grammar}
\begin{footnotesize}

\subsubsection{Productions describing the structure of the sentences}

\vspace{0.5cm}
CONSTRUCTION $\rightarrow$ AGREEMENT

\vspace{0.5cm}
\noindent
AGREEMENT $\rightarrow$ MAIN-CLAUSE

\noindent
AGREEMENT $\rightarrow$ COMPLETIVE-CLAUSE

\noindent
AGREEMENT $\rightarrow$ RELATIVE-CLAUSE

\vspace{0.5cm}

\noindent
MAIN-CLAUSE $\rightarrow$ 
SUBJNP(Num) ATTRACTORS VERB(Num) \{PP-TMP, PP-MANNER\}

\noindent
COMPLETIVE-CLAUSE $\rightarrow$ 
SUBJNP(Num) VERB(Num) MAIN-CLAUSE

\noindent
RELATIVE-CLAUSE $\rightarrow$ 
SUBJNP(Num) ATTRACTORS RELCLAUSE VERB(Num) \{PP-TMP, PP-MANNER\}

\vspace{0.5cm}

\noindent
SUBJNP(Num) $\rightarrow$ Det(Num) N(Num)

\noindent
SUBJNP(Num) $\rightarrow$  N(Num)

\noindent
VERB(Num) $\rightarrow$ verb

\noindent
PP-TMP $\rightarrow$ \{ a certain number of fixed expressions\}

\noindent
PP-MNR $\rightarrow$ \{ a certain number of fixed expressions\}

\noindent
RELCLAUSE $\rightarrow$ \{ a certain number of fixed expressions\}

\noindent
ATTRACTORS  $\rightarrow$ PP

\noindent
ATTRACTORS  $\rightarrow$ PP ATTRACTORS

\noindent
PP $\rightarrow$ P NP

\noindent
P $\rightarrow$ \{ prepositions \}

\noindent
NP $\rightarrow$  \{ a certain number of fixed expressions\}

\subsubsection{Rules that apply to subject NPs, verb and NPs inside attractors.}

\noindent
ALTERNATE:  for each rule object, picks an attribute, e.g. number, and loops over all the possible $k$ values with increments of $k^n$, for $0 \le k \le n$.

In our case, for example, it samples without replacement all NPs and verb, and loops over the values of the attribute $number$  with increasing increments every other one, every second one, every four ones.

\vspace{0.2cm}
\noindent
PROGRESSION: applies to attractors and it creates a progression 0,1,2,3,4 ...


\end{footnotesize}	

\newpage
\subsection{Original French examples}	
\label{original-french-examples}

We generated sentences according to the rules with the adapted items from \citet{Franck-ea2002}. We also created three versions of sentences: noun phrases in the main clause (\cref{fig:matrix}) or  embedded in a completive (\cref{fig:completive}) or  a relative clause (\cref{fig:relative}).

\begin{figure}[!ht]
\small
\centering
\subfloat[] [Main clause \label{fig:matrix}]{ 

\begin{tabular}{llll} 
\hline
1 L’ordinateur    & avec le programme    & &  \textbf{est} en panne. \\
2 Les ordinateurs & avec le programme    & & \textbf{sont} en panne. \\
3 L’ordinateur    & avec les programmes & &  \textbf{est} en panne. \\
4 Les ordinateurs & avec les programmes  & & \textbf{sont} en panne. \\
5 L’ordinateur    & avec le programme    & de l’expérience&  \textbf{est} en panne.	\\
6 Les ordinateurs & avec le programme    & de l’expérience&  \textbf{sont} en panne. \\	
7 L’ordinateur    & avec les programmes  & de l’expérience& \textbf{est} en panne. \\
8?Les ordinateurs & avec les programmes  & de l’expérience & \textbf{sont} en panne.\\
\hline 
\end{tabular} 
 } 

\subfloat[][Completive clause \label{fig:completive}]{

\begin{tabular}{llll} 
\hline
1 Jean suppose que l’ordinateur    & avec le programme    & &  \textbf{est} en panne. \\
2 Jean suppose que les ordinateurs & avec le programme    & & \textbf{sont} en panne. \\
3 Jean suppose que l’ordinateur    & avec les programmes & &  \textbf{est} en panne. \\
4 Jean suppose que les ordinateurs & avec les programmes  & & \textbf{sont} en panne. \\
5 Jean suppose que l’ordinateur    & avec le programme    & de l’expérience&  \textbf{est} en panne.	\\
6 Jean suppose que les ordinateurs & avec le programme    & de l’expérience&  \textbf{sont} en panne. \\	
7 Jean suppose que l’ordinateur    & avec les programmes  & de l’expérience& \textbf{est} en panne. \\
8?Jean suppose que les ordinateurs & avec les programmes  & de l’expérience & \textbf{sont} en panne.\\
\hline 
\end{tabular} 
}

\subfloat[][Relative clause \label{fig:relative}]{

\begin{tabular}{llll} 
\hline
1 L’ordinateur    & avec le programme    & &  dont Jean se servait \textbf{est} en panne. \\
2 Les ordinateurs & avec le programme    & &  dont Jean se servait \textbf{sont} en panne. \\
3 L’ordinateur    & avec les programmes  & &  dont Jean se servait \textbf{est} en panne. \\
4 Les ordinateurs & avec les programmes  & &  dont Jean se servait \textbf{sont} en panne. \\
5 L’ordinateur    & avec le programme    & de l’expérience&  dont Jean se servait \textbf{est} en panne.	\\
6 Les ordinateurs & avec le programme    & de l’expérience&  dont Jean se servait \textbf{sont} en panne. \\	
7 L’ordinateur    & avec les programmes  & de l’expérience&  dont Jean se servait \textbf{est} en panne. \\
8?Les ordinateurs & avec les programmes  & de l’expérience & dont Jean se servait \textbf{sont} en panne.\\
\hline 
\end{tabular} 

}
\caption{Sentences generated with items (N1= ordinateur, N2=programme, N3= expérience). There are three types of context: main clause, completive clause, and relative clause.}   \label{fig:context}
\end{figure}	

\begin{figure*}[h] 
\small
\centering
\begin{tabular}{lp{0.55\linewidth}p{0.40\linewidth}}\hline
&Answers & Error type\\\hline
1 &L’ ordinateur  avec le programme et l'expérience est en panne. & Coord: the last noun appears in a coordination phrase\\
2 &\textbf{Les ordinateurs  avec les programmes de l’expérience  sont en panne.}  & Correct: Correct answer\\
3 &L’ ordinateur  avec le programme est en panne.  & WNA: Wrong number of attractors \\
4 &L’ ordinateur  avec le programme de l’expérience sont en panne. & AE: Agreement error\\
5 &Les ordinateurs  avec le programme de l’expérience sont en panne. &   Alter N1: wrong number of the first NP attractor\\
6 &Les ordinateurs  avec les programmes des expériences  sont en panne. & Alter N2: wrong number of the second NP\\\hline 
\end{tabular}
\caption{Different conditions of the answer matrix for the main clause template.}
\label{answers}
\end{figure*}

\newpage
\subsection{Masking}	
\label{masking-examples}

  \begin{itemize}
\item For the main clause, we masked the first noun, and generated the five most probable nouns. We applied the same procedure for the second noun. 
    
   a. Les MASK avec le programme  de l’expérience \textbf{sont}  en panne. \label{mask:maina}
   
   b. Les ordinateurs avec le MASK  de l’expérience \textbf{sont}  en panne.  \label{mask:mainb}
     
\item For the completive clause, we mask the head  noun in the subject and main verb, as well as first noun   and second noun of the embedded clause. 
    
    a. MASK suppose que les ordinateurs avec le programme de l’expérience \textbf{sont}  en panne.\label{mask:coma}
    
   b. Jean MASK  que les ordinateurs avec le MASK  de l’expérience \textbf{sont}  en panne. \label{mask:comb}
   
     c. Jean suppose que les  MASK  avec le programme  de l’expérience \textbf{sont}  en panne. \label{mask:comc}
     
   d. Jean suppose que les ordinateurs avec le  MASK    de l’expérience \textbf{sont}  en panne. \label{mask:comd}

\item For the  relative clause, we mask the head  noun   and main verb of the relative clause, as well as the first noun  and second noun of the subject noun phrase of the main clause.
    
       a.  Les MASK avec le programme  de l’expérience dont Jean se servait \textbf{sont}  en panne. \label{mask:rela}
       
   b.  Les ordinateurs avec le MASK  de l’expérience dont Jean se servait \textbf{sont}  en panne. \label{mask:relb}
   
   c.  Les ordinateurs avec le programme    de l’expérience dont MASK  se servait \textbf{sont}  en panne. \label{mask:relc}
   
   d.  Les ordinateurs avec le programme    de l’expérience dont Jean se  MASK \textbf{sont}  en panne.  \label{mask:reld}
   \end{itemize}

\end{document}